\documentclass[runningheads,a4paper]{llncs}

\usepackage{amssymb,amsmath}
\usepackage{color}
\usepackage{graphicx}
\usepackage{url}
\usepackage{ulem}
\newcommand{\keywords}[1]{\par\addvspace\baselineskip
\noindent\keywordname\enspace\ignorespaces#1}

\definecolor{mygrey}{rgb}{0.32549,0.3451,0.37255}
\definecolor{myorange}{rgb}{1.,0.52157,0.23922}
\definecolor{myblue}{rgb}{0.011765,0.38039,0.75294}

\newcommand{\ve}[1]{\mathbf{#1}}

\begin{document}

\normalem
\mainmatter  % start of an individual contribution

\title{Algorithmic Composition of Melodies with Deep Recurrent Neural Networks}

\titlerunning{Algorithmic Composition of Melodies with Deep Recurrent Neural Networks}

\author{Florian Colombo, Samuel P. Muscinelli, Alexander Seeholzer, Johanni Brea  \and Wulfram Gerstner}
%
% if the names of the authors are too long for the running head, please use the format: AuthorA et al.
\authorrunning{Colombo et al.}

\institute{Laboratory of Computational Neurosciences. Brain Mind Institute. \'Ecole Polytechnique F\'ed\'erale de Lausanne\\ \email{florian.colombo@epfl.ch}}


\maketitle

\begin{abstract}
A big challenge in algorithmic composition is to devise a model that
is both easily trainable and able to reproduce the long-range temporal
dependencies typical of music.  Here we investigate how artificial neural
networks can be trained on a large corpus of melodies and turned into
automated music composers able to generate new melodies coherent with the style
they have been trained on. We employ gated recurrent unit networks that have
been shown to be particularly efficient in learning complex sequential
activations with arbitrary long time lags.  Our model
processes rhythm and melody in parallel while modeling the relation between
these two features. Using such an approach, we were able to generate
interesting complete melodies or suggest possible continuations of a melody
fragment that is coherent with the characteristics of the fragment itself.

\end{abstract}

\section{Introduction}

The algorithmic formalization of musical creativity and composition, foreseen 
as early as the 19th century \cite{Lovelace1843}, has come to fruition in the recent
decades with the advent of modern computer algorithms \cite{Fernandez2013}.

Formally, a melody can be seen as a sample from a potentially very sophisticated
probability distribution over sequences of notes
\cite{Fernandez2013,Jones1981,Ames1989,Papadopoulos1999}.  For monophonic music,
probability distributions could be given by Markov chains, where the probability
of the next note depends only on the current note and the $k$ last notes
\cite{Ames1989}. Markov chain models, however, do not capture the long-range
temporal structure inherent in music.
For example, even a simple melody such as \emph{Brother John} is structured in
four patterns, each repeated twice, with the first and last ones starting with
the same notes (see Fig. \ref{fig:representation}). Taking only the last few notes into account is thus not enough
to produce the sequence -- rather does the progression on the long timescale of
bars dictate the sequences of notes. 

%REVIEW p.1, Include the melody of Brother John segmented in the 4 parts mentioned.

Such rich temporal structure can be captured by models that rely on recurrent neural
networks (RNN). Particularly well suited to capture these long-range temporal dependencies are
models based on long short-term memory (LSTM) units \cite{Hochreiter1998} and
variants thereof \cite{Gers2000,Chung2014,Cho2014}.
Thanks to automatic differentiation, parallel use of GPUs, and software packages
like \verb|theano| \cite{Bergstra2010} or \verb|torch| \cite{Collobert2011}, 
it has become possible to easily fit such models to large
scale data and obtain impressive results on tasks like text translation
\cite{Sutskever2014}, speech recognition \cite{Graves2013speech} and text or
code generation \cite{Graves2013,Karpathy2015}. 
For algorithmic composition of monophonic music, it has been observed that RNN
models based on LSTM units \cite{Eck2002,Franklin2004} can capture 
long-term temporal dependencies far better than RNNs with simple units 
\cite{Todd1989,Mozer1994}.

Closest to our approach is the work of Eck and Schmidhuber \cite{Eck2002} and
Franklin \cite{Franklin2004}. Eck and Schmidhuber used an LSTM-RNN to generate a
fixed Blues chord progression together with well-fitting improvised melodies. To
model polyphonic music they discretized time into bins of equal duration, which
has the disadvantage of using the same representation for one long note, e.g. a
half note, and repeated notes of the same pitch, e.g. two quarter notes.
Franklin experimented with different representations of pitch and duration and
used an LSTM-RNN to reproduce a single long melody. These studies showed that
LSTM-RNN's are well suited to capture long-range temporal dependencies in music,
but they did not demonstrate that these models can autonomously generate
musically convincing novel melodies of a given style extracted from large datasets.

%REVIEW The 3rd paragraph of the Discussion section would be better suited for the literature review presented in the Introduction, given the level of details it provides about the system developed by Eck and Schmidhuber. + The paper lacks a proper review of the AI systems able to generate melodies automatically. A few works are mentioned, but these only scratch the surface of what has been done in the last few decades. The survey by Fernandez and Vico (2013) can be used to find additional information about the topic. It would be interesting to discuss other methods used for automatic melody generation (e.g., Genetic Algorithms, Generative Grammars) comparing them with the proposed system. }

Here, we present a deep (multi-layer) model for
algorithmic composition of monophonic melodies, based on RNN with gated recurrent units (GRU).
We selected GRUs because they are simpler than
LSTM units, but at least equally well suited to capture long-range
temporal dependencies when used in RNNs (see \cite{Chung2014,Jozefowicz2015} for comparison studies).
Our model represents monophonic music as a sequence of notes, where each note is
given by its duration and pitch.
Contrary to earlier studies, which evaluated LSTM based models of music 
only on small and artificial datasets \cite{Eck2002,Franklin2004}, 
we train our model on a large dataset of Irish folk songs. 
We apply the trained model to two different tasks: proposing
convincing continuations of a melody and composing entire new songs.

%REVIEW Explain why you preferred GRU to LSTM.
%REVIEW What is the actual difference between GRU and LSTM?
\section{Methods}
%Proposed method: This most important part of the paper. Give all important details, assumptions and working of proposed method without going into minute detail of each and every step.
%

To devise a statistical model that is able to complete and autonomously produce
melodies, we train multi-layer RNNs on a large corpus of Irish folk songs.
In the training phase, the network parameters are updated in order to accurately
predict each upcoming note given the previously presented notes of songs in a
training set (Fig.~\ref{fig:network}A). The model, once trained, can then be used
to generate the upcoming notes itself, consequently producing whole new musical sequences.

After introducing our representation of melodies, we give a short introduction
to recurrent neural networks and
present the model and the training modalities.
Finally, we explain how the model is used for algorithmic
composition.

\subsection{Music Representation}

\begin{figure}[t]
\centering
\includegraphics[width=1.0\textwidth]{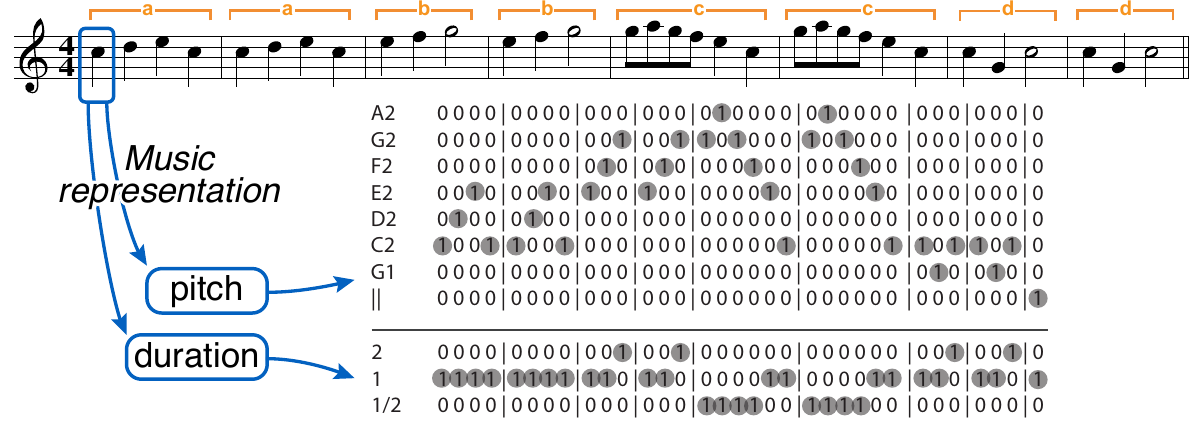} 
\caption{\textbf{Representation of a simple melody}: The nursery rhyme \emph{Brother John} in symbolic music notation, represented by pitch and duration matrices. Each column represents a single note. The four patterns composing the song are labeled by a, b, c and d.}
\label{fig:representation}
\end{figure}

Melodies are represented as sequences of notes, where each note is a combination
of its pitch and duration. As such, each note $n$ can be represented by two
corresponding one-hot vectors (with only one entry of $1$, and $0$ for all
others) of pitch $\mathbf{p}[n]$ and duration $\mathbf{d}[n]$ . These vectors
encode pitches and durations present in the
training set. In addition, we include ``song ending''
and ``silence'' features as supplementary dimensions of the pitch vector. Any
song can thus be mapped to a matrix of pitches and a second matrix of durations.

To reduce dictionary sizes as well as redundancies in representations, we chose
to normalize both melody and duration. Melodies are normalized by transposing
every song into C Major/A minor. Durations are normalized as relative to the
most common duration in each song: if, for example, the most common duration in a song is the
quarter note, we would represent eighth notes as ``1/2''. For an example of a melody in this representation see Fig. \ref{fig:representation}.

%REVIEW The description of the musical representation employed by the system is a tad opaque. To clarify the musical encoding, include - as a figure? - the musical representation of a short melody.

%After this normalization, the pitch value is absolute with each element of the pitch vector associated with a single pitch. 
%Both features are represented as random variables in a 1-of-K representation (Equation \ref{eq:1ofK}). 
%\begin{equation}
%\label{eq:1ofK}
%\mathbf{x} = \left[x^1\dots x^k\dots x^D \right]^\intercal \quad \sum_k x^k = 1 \quad x^k \in \{0,1\} \;  . \\
%\end{equation}
%\sam{To have a coherent representation of duration across melodies, we normalize durations in the following way.} Each element of the duration binary vector is associated with a single duration, which value is relative to the most common duration in each melody. For example, if the most common duration in a song is the quarter note, each quarter note is associated to the element corresponding to a 1 in the duration vector. Consequently, eight notes will be associated to the 1/2 element and eight note triplets to 1/3. In a music corpus with these three elements only, the duration vector for each quarter note is $[0,0,1]$, $[0,1,0]$ for an eight note and $[1,0,0]$ for an eight note triplet. 
%This representation keeps all the information needed to retrieve the original melodies. 

\subsection{A Brief Introduction to Recurrent Neural Networks}
Artificial neural networks have a long history in machine learning, artificial
intelligence and cognitive sciences (see \cite{deeplearningbook} for a textbook,
\cite{Hinton2015} for recent advances, \cite{Schmidhuber2014} for an in-depth
historical overview). Here we give a brief introduction for readers unfamiliar
with the topic.  

Artificial neural networks are non-linear functions $\ve{y} = f_\ve{w}(\ve{x})$,
where input $\ve x$, output $\ve y$ and the parameters (weights) $\ve w$ can be
elements of a high-dimensional space. A simple example of an artificial neural
network with 2-dimensional in- and outputs is given by $y_1 = \tanh(w_{11}x_1 +
w_{12} x_2)$ and $y_2 = \tanh(w_{21}x_1 + w_{22}x_2)$, which we write in short
as $\ve y = \tanh(\ve w \ve x)$.  Characteristic of artificial neural networks
is that the building blocks consist of a non-linear function $\sigma$ (like
$\tanh$) applied to a linear function $w_{11}x_1 + w_{12} x_2$, which is an
abstraction of the operation of biological neurons. If these building blocks
are nested, e.g. ${\ve y = \sigma_3\Big(\ve w_3 \sigma_{2}\big(\ve w_{2}
\sigma_1(\ve w_1 \ve x)\big)\Big)}$, one speaks of multi-layer (or deep) neural
networks, with layer-specific weights ($\ve w_1, \ve w_2, \ldots$) and
non-linearities ($\sigma_1,\sigma_2,\ldots$). 

Deep neural networks are of
particular interest for the approximation of high-dimensional and non-linear
functions. For example, the function of recognizing object $i$ in photos can be
approximated by adjusting the weights such that output $y_i$ is 1 if and only if
an image $\ve x$ of object $i$ is given \cite{Hinton2015}.
Formally, the weights can be adjusted to minimize a
cost function, like the averaged square loss between target and output ${\mathcal
L(\ve w) = \frac1S\sum_{s=1}^S\big(\ve y_s - f_{\ve w}(\ve x_s)\big)^2}$ for some known
input-output pairs $(\ve x_s,\ve y_s)$. Since artificial neural networks are
differentiable in the parameters $\ve w$, this cost function is also
differentiable and the parameters can be adjusted by changing them in direction
of the gradient of the cost function $\Delta \ve w \propto \nabla_{\ve w}\mathcal
L(\ve w)$. 

In recurrent neural networks (RNN) the inputs and outputs are sequences of
arbitrary length and dimension. A simple example of a recurrent neural network
with one hidden layer is given by $\ve h[n] = \sigma(\ve w_{xh}\ve x[n] + \ve
w_{hh}\ve h[n-1])$ and $\ve y[n] = \sigma(\ve w_{hy} \ve h[n])$, where $\ve
x[n],\, \ve h[n],\, \ve y[n]$ is the $n$-th element of the input, hidden, output
sequence, respectively. This network is recurrent, since each hidden state $\ve
h[n]$ depends on the previous hidden state $\ve h[n-1]$ and, therefore, on all
previous input elements  $\ve x[1], \ve x[2],\ldots, \ve x[n]$.  While these
recurrent neural networks can in principle capture long-range temporal dependencies, 
they are difficult to fit to data by gradient descent, since
the gradient involves the recurrent weights $w_{hh}$ raised to high powers, which
vanishes or explodes depending on the largest eigenvalue of $w_{hh}$
\cite{Hochreiter1998}. This problem can be avoided by a
reparametrization of the recurrent neural network (LSTM \cite{Hochreiter1998},
GRU \cite{Cho2014}, other variants \cite{Jozefowicz2015}). In Equation
\ref{eq:hGRU} we give the update equations for the GRU used in this study. 

\subsection{Model}

\begin{figure}[t]
\centering
\includegraphics[width=1.0\textwidth]{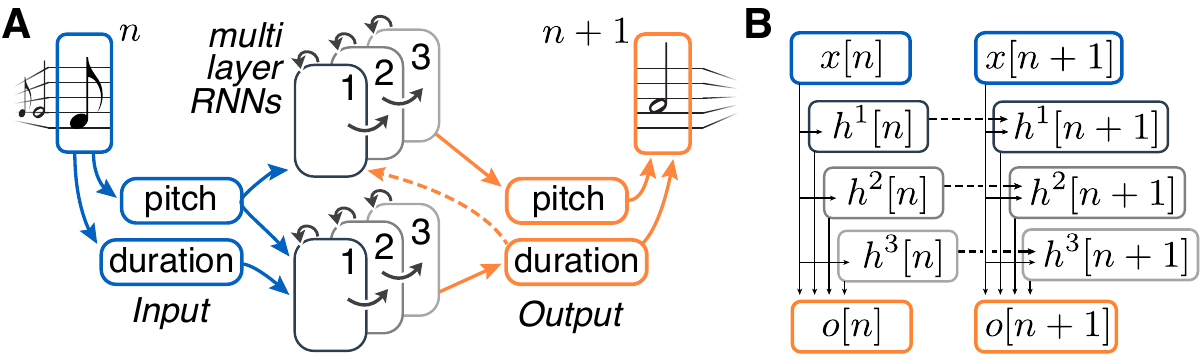} 
\caption{\textbf{Network architecture and connectivity}: \textbf{A} Each note in
a melody is separated into pitch and duration components, which are iteratively
fed into two multilayer RNNs as inputs. Network outputs give probability
distributions for pitch and duration of the next note. The rhythm network receives the current duration and pitch while the melody network
receives the current pitch and the upcoming duration as input (dashed line).\textbf{B} The input layer $x$ projects to all hidden layers $h$, as well as the output layer $o$. The hidden layers $h$ project recurrently between notes (dashed
lines) as well as feeding forward to all higher hidden layers and the output layer.}
\label{fig:network}
\end{figure}

To model distributions of pitch and duration, we use two separate multi-layer RNNs
(see Fig. \ref{fig:network}A) referred to as the \emph{rhythm} and \emph{melody} networks, respectively. The numbers of output units are equal
to the dictionary sizes of our music representation. The three hidden layers of both the 
pitch and duration RNNs are composed of 128 GRUs \cite{Cho2014} each, and are connected as shown in
Fig. \ref{fig:network}B. The model was implemented using the \verb|theano| library \cite{Bergstra2010}.

%REVIEW Given the interdisciplinary context of the conference, a short introduction to RNN would improve readability a lot. + A background section, introducing relevant elements such as RNNs, GRU, and gates, would make the paper self-contained.
 
For each note $n$ in a musical sequence, the the
duration vector $\mathbf{d}[n]$ of this note is presented to the rhythm network, while the melody network receives
both the pitch vector $\mathbf{p}[n]$ as well as the duration vector of the
\emph{upcoming} note $\mathbf{d}[n+1]$ as inputs. 
Each time a note is fed to the model, all internal states are updated and the rhythm network output gives a distribution over the possible upcoming durations $Pr(\mathbf{d}[n+1]|\mathbf{d}[n])$. In the same way, the melody network output gives a distribution over possible upcoming pitches $Pr(\mathbf{p}[n+1]|\mathbf{p}[n], \mathbf{d}[n+1])$. 

The update equations for the vector of layer activations $\mathbf{h}^i[n]$,
update gates $\mathbf{z}^i[n]$ and reset gates $\mathbf{r}^i[n]$ at
note $n$ for layer $i\in\{1,2,3\}$ are given by 
\begin{align}
\label{eq:hGRU}
\mathbf{h}^i[n] &= \mathbf{z}^i[n] \odot \mathbf{h}^i[n-1] +
(\mathbf{1}-\mathbf{z}^i[n]) \odot \widetilde{\mathbf{h}}^i[n] \; ,
\\
\label{eq:htildaGRU}
\widetilde{\mathbf{h}}^i[n] &= \tanh \bigg( \ve w_{y^ih^i} \mathbf{y}^i[n] +
\mathbf{r}^i[n] \odot \ve w_{h^ih^i} \mathbf{h}^i[n-1] \bigg)\; ,
\\\label{eq:zGRU}
\mathbf{z}^i[n] &= \sigma \bigg(\ve w_{y^iz^i}\mathbf{y}^i[n] +  \ve w_{h^iz^i}\mathbf{h}^i[n-1] + \mathbf{b}^i_z\bigg) \; ,
\\\label{eq:rGRU}
\mathbf{r}^i[n] &= \sigma \bigg(\ve w_{y^ir^i}\mathbf{y}^i[n] +
\ve w_{h^ir^i}\mathbf{h}^i[n-1] + \mathbf{b}^i_r\bigg) \; ,
\end{align}
where 
$\sigma(x)=(1+\exp(x))^{-1}$ is the logistic sigmoid function, $\odot$ denotes the
element-wise product and $\mathbf{y}^i$
is the feed-forward input to layer $i$, which consists
of both the global inputs $\mathbf{x}[n]$ as well as hidden layer
activations $\mathbf{h}^{j<i}[n]$ (see Fig.~\ref{fig:network}B). The update equation for the output unit $i$ activation $\mathbf{o}_i[n]$ at note $n$ is 
\begin{equation}
\mathbf{o}_j[n] = \Theta \big(\ve w_{y^oo}\mathbf{y}^o[n] + \mathbf{b}^o\big)_j\;  ,
%\theta(\mathbf{x})_i &= \frac{e^{x_i}}{\sum_k e^{x_k} }
\label{eq:o}
\end{equation}
where $\mathbf{y}^o$ is the feed-forward input of the output layer and $\Theta(\mathbf{x})_j = \frac{e^{x_j}}{\sum_k e^{x_k} }$ is the Softmax function. The Softmax normalization ensures that the values of the output units sum to one, which allows 
us to interpret the output of the two RNNs as probability distributions over pitches and over durations.
For example, the probability of pitch $j$ of an upcoming note is then given
by
\begin{equation}
Pr\big(\mathbf{p}_j[n+1] = 1\mid \mbox{previous notes and } {\theta}) =
	\mathbf{o}_j^{Melody}[n]\; ,
\end{equation}
where the conditioning on previous notes and the model parameters $\theta$
highlights that the network output $\mathbf{o}[n]$ depends on them.

%From Equation \ref{eq:hGRU} one can see that the GRU carries a trace of an eventin its activity as long as its update and reset gates are non-zero. The network can learn when exactly to open or close these gates, which helps GRUs to perform well on learning long-range temporal dependencies in sequences \cite{Cho2014}.

\subsection{Training \& Melody Generation}

During training, the log-likelihood of model parameters $\theta$ for the
rhythm and melody networks are separately optimized by stochastic
gradient ascent with adaptive learning rate \cite{Kingma2014} ($\alpha=10^{-3}$,
$\beta_1=0.9$,  $\beta_2=0.999$ and $\epsilon=10^{-8}$). The log-likelihood of
model parameters $\theta$ given the training songs is given by
\begin{equation}
\label{eq:nll}
\mathcal{L}(\theta) =
\frac{1}{S}\sum_{s=1}^S\frac{1}{N_s-1}\sum_{n=1}^{N_s-1}\log\Big(Pr\big(\mathbf{x}^s_j[n+1] = 1
\mid \mbox{previous notes and } {\theta}\big)\Big)\; ,
\end{equation}
where $S$ is the total number of songs, $N_s$ the length of  song
$s$ and  $\mathbf{x}^s[n]$ is the duration vector $\mathbf{d}^s[n]$ for the rhythm network and the pitch vector $\mathbf{p}^s[n]$ for the melody network. The trainable model parameters $\theta$ are the connection matrices
$\ve w_{ab}$ for $a\in\{y^i,h^i\},\ b\in\{h^i,z^i,r^i\}$ and $\ve w_{y^oo}$, the gate
biases $\mathbf{b}_z^i$ and $\mathbf{b}_r^i$, the output unit biases
$\mathbf{b}^o$ and the initial state of the hidden units
$\mathbf{h}^i[0]$.

Networks are trained on 80\% of songs of the musical corpus and tested on the
remaining 20\%. A training epoch consists of optimizing model parameters on each
of randomly selected 200 melodies from the training set, where parameters are
updated after each song. After each epoch, the model performances are evaluated
on a random sample of 200 melodies from the testing and the training set. The
parameters that minimize the likelihood on unseen data from the testing set are
saved and used as final parameters.

Melody generation is achieved by closing the output-input loop of Eqs.
\ref{eq:hGRU}-\ref{eq:o}. In each time step a duration and a pitch are sampled
from the respective probability distributions and used as inputs in the next
time step.

%In the song continuation task, the beginning of a sequence $\mathbf{x}^{seed}[1:n_0]$ taken from the testing dataset, which we call \textit{seed}, is fed to the model. It effectively will bring the model dynamic in a state that eventually allows the trained model to produce a correct prediction of the next note given the last input $\mathbf{x}^{seed}[n_0]$ and the model internal representation of the sequence history.
%The predicted input of the next time step $\mathbf{\hat{x}}[n_0+1]$ can be obtained by sampling the model outputs $Pr(\mathbf{d}[n_0+1])$ and $Pr(\mathbf{p}[n_0+1])$.
%Sampling $\mathbf{\hat{d}}[t+1]$ ($\mathbf{\hat{p}}[n+1]$ respectively), is performed by drawing from a categorical distribution with parameters $Pr(\mathbf{d}[n+1])$ ($Pr(\mathbf{p}[n+1])$).
%Using the predicted output $\mathbf{\hat{x}}[n_0+1]$ as the new input $\mathbf{x}[n_0+1]$, a process that can be thought as \emph{machine dreaming}, allows to iteratively add an artificially generated note to the seed. This process can then be carried on to generate an infinite number of possible sequels. 
%
%In the autonomous song generation task, we let the model compose from the beginning by only feeding the first two notes to start its generative process. Melody generation is carried on until the song ending tag is sampled from the melody network.

\section{Results}
%Experimental data and results: In this section give brief information about the data sets used and their properties. Results of proposed method after applying different data set to it must me added. Comparing with other methods can optional but adding them will definitely strengthen acceptability of the paper.
%
\begin{figure}[t]
\centering
\includegraphics[width=1.0\textwidth]{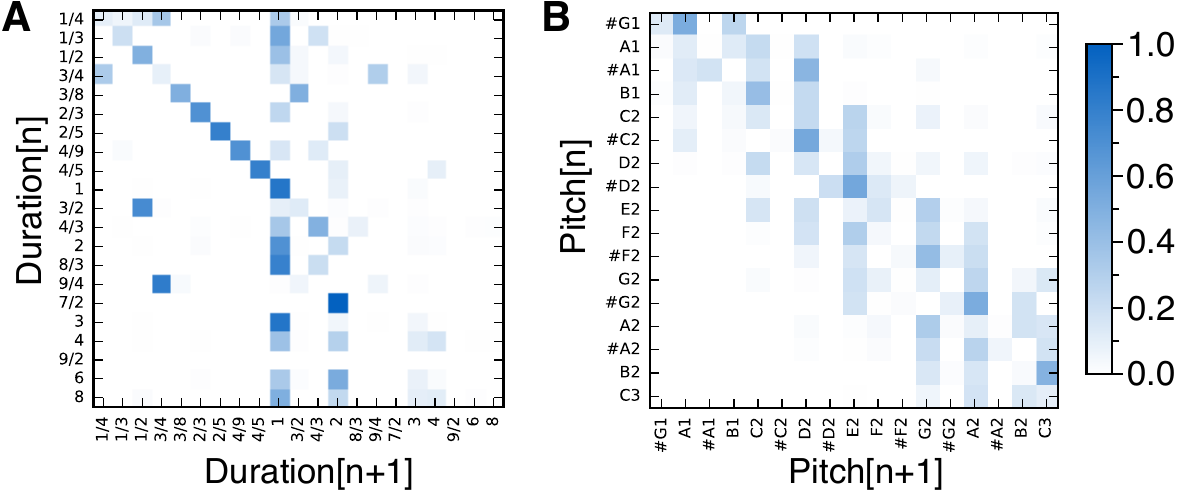} 
\caption{\textbf{Transition probabilies in the music corpus}: \textbf{A} Probability of transitions from durations at note $n$ to note $n+1$ color graded from zero (white) to one (blue). \textbf{B} Same as in panel A, for transitions of a selected range of pitches.}
\label{fig:markov}
\end{figure}

\subsection{Music Corpus}

To automatically convert music corpora to our representation, we developed a
toolbox that parses symbolic music files written in the abc
notation\footnote{http://abcnotation.com/} and converts them into our music representation. The results presented here are based on the Irish
music corpus of Henrik Norbeck\footnote{http://www.norbeck.nu/abc/}. It contains
2158 Irish tunes with an averaged length of $136\pm84$ notes. The basic
statistics of pitch and duration transitions are shown in Fig. \ref{fig:markov}. Due to our
normalization procedure, the most common duration is 1 and there is consequently
a high probability of transition to the unitary duration from any state.
Otherwise, shorter durations are mostly followed by the same or
a complementary value, e.g. ``3/2'' followed by ``1/2''. 
The most common pitches belong to the diatonic C Major
scale. Transition from natural B, respectively G sharp, have high probability of
ending in the closest C, respectively A, a property inherent to western music in
order to resolve a melody. 

The analysis of the relation between the pitch and duration features revealed that they are dependent, as expected from music theory. Therefore, we explicitly modeled the distribution over upcoming pitches as depending on the upcoming duration (dashed line in Fig. \ref{fig:network}A), effectively splitting the joint distribution over note duration and pitch into conditional probabilities.

\begin{figure}[ht!]
\centering
\includegraphics[width=1.0\linewidth]{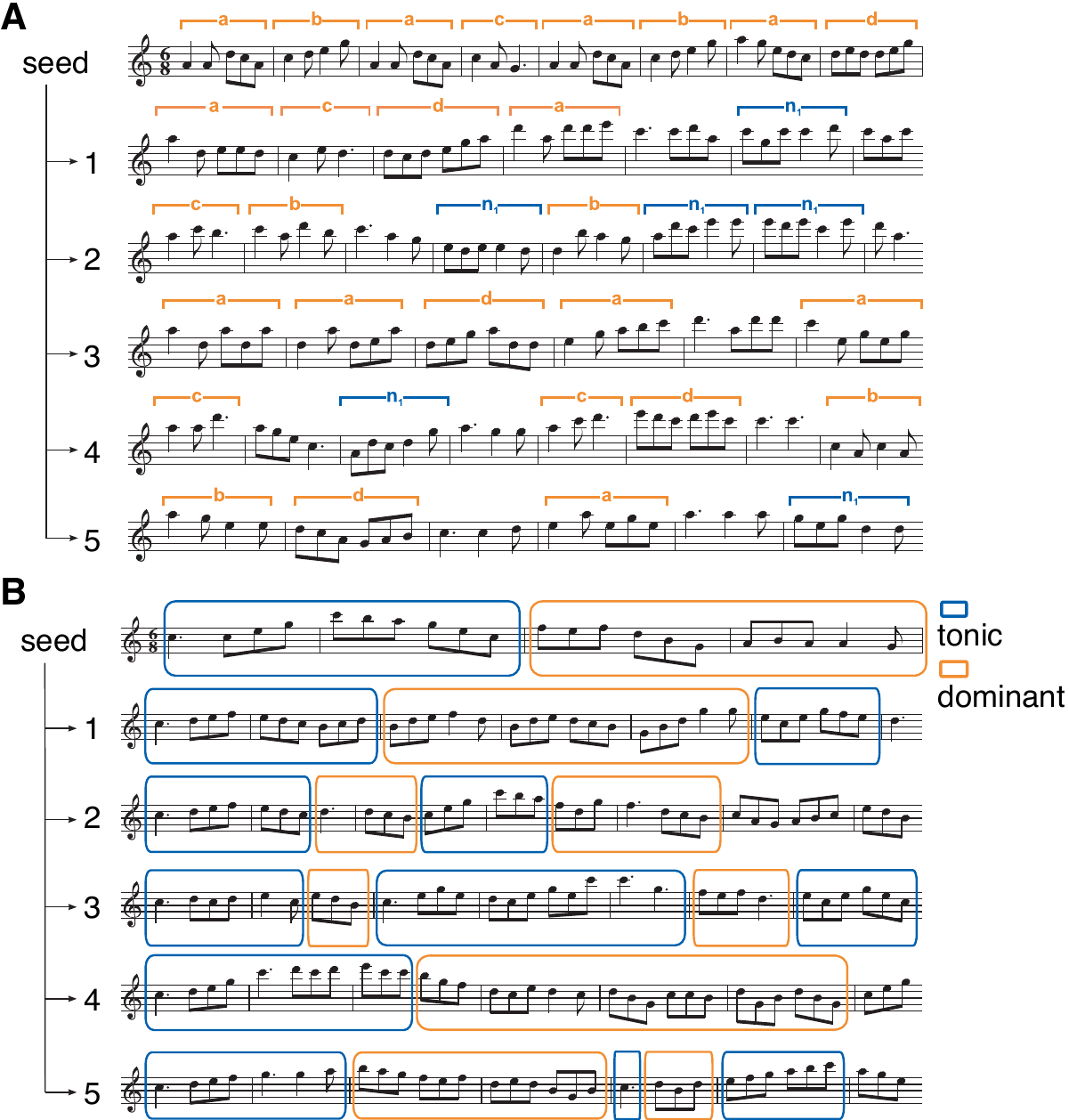}
\caption{\textbf{Example of melody continuation and analysis}: \textbf{A} The first line (seed) is presented as input to the model. The next five lines are five different possible continuations generated by our model. Rhythmical patterns present in the seed are labeled a, b, c and d. The label n$_1$ points at a new rhythmical pattern generated by the model. Unlabeled bars show other novel rhythmical patterns
that appear only once. \textbf{B} A second example of song continuation, with analysis of tonal areas.
}
\label{fig:CAAC}
\end{figure}

\subsection{Song Continuation}

%Here, our model is trained in order to optimize the likelihood of the model
%parameters given unseen music example from the same music corpus. We expect that
%the RNNs will learn to extract from observed examples the relation between notes
%both in term of rhythm and melody such that it could make correct predictions on
%the upcoming notes given the network internal representation of the sequence
%history. 

%After training, the rhythm and melody networks are evaluated on the testing data. The performances are evaluated in term of average negative log-likelihood (Equation \ref{eq:nll}) and average prediction accuracy on each song. A prediction is considered correct when the unit associated to the true upcoming duration or melody is the most active. A song predicted with 100\% accuracy could therefore lead to different outcomes through the sampling mechanisms as other units could also be active. 

%The accuracy Fig. \ref{fig:CAACtraining} shows the evolution during training on the irish dataset of these two quantities. It is observed that the performances on unseen data from the validation set increases during training until the model starts to be more specific to the training data and consequently less general. Therefore, for compter-aided algorithmic composition, we save the model parameters that leads to better performances on the validation set. 

To study song continuations, we present as input to the trained model the
beginnings of previously unseen songs (the \emph{seed}) and observe several continuations that 
our model produces (see Fig. \ref{fig:CAAC}). 
From a rhythmical point of view, it is interesting
to notice that, even though the model had no notion of bars implemented, the
metric structure was preserved in the generated continuations. 
Analyzing the continuations in Fig. \ref{fig:CAAC}A, we see that the 
A pentatonic scale that characterizes the seed is maintained everywhere except 
for continuation number 2.  This is noteworthy,
since the model is trained on a dataset that does not only contain music based
on the pentatonic scale.  
Moreover, the rhythmic patterns of the seed are largely maintained in the
continuations.  Rhythmical patterns that are extraneous to the seed are also generated. 
For example, the pattern $n_1$, which  is the
inversion of pattern $a$, can be observed several times.  Importantly, the
alternating structure of the seed ($abacabad$) is not present in the generated
continuations. This indicates that the model is not able to capture this level
of hierarchy.
While less interesting than the first example from the
rhythmical point of view, the seed presented in Fig. \ref{fig:CAAC}B, is clearly
divided in a tonic area and a dominant area.  Different continuations are
coherent in the sense that they also alternate between these two areas, while
avoiding the subdominant area almost everywhere.

\subsection{Autonomous Song Generation}

\begin{figure}[t]
\centering
%\flushleft{A}
%\includegraphics[width=\textwidth]{./Figures/AMC/AMC1_original.pdf}
%\flushleft{B}
\includegraphics[width=.8\textwidth]{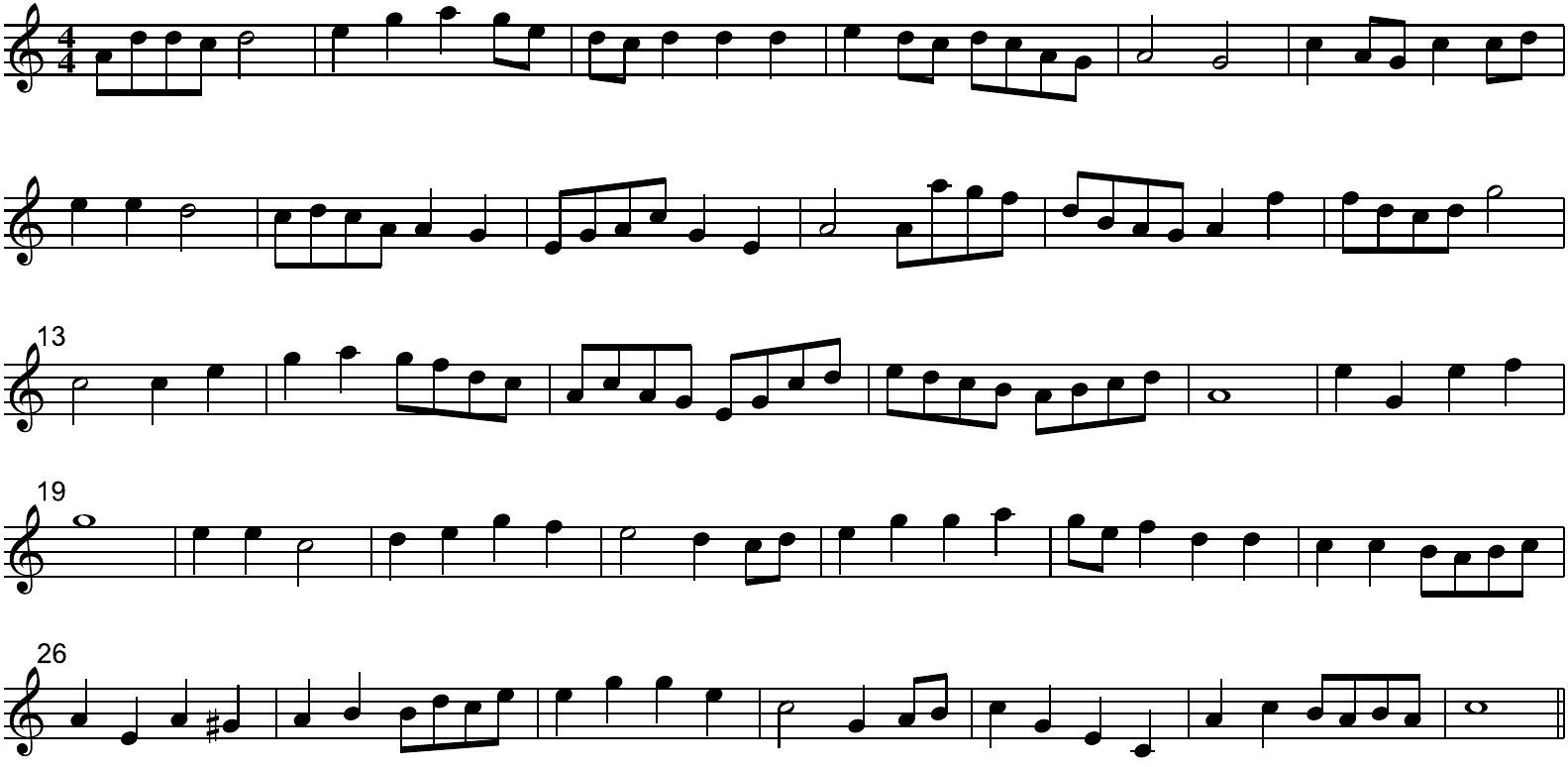}
\caption{\textbf{Example of an autonomously generated Irish tune}: See text for details. In this example a coherent temporary modulation to A minor was generated (bar 26).
Worth noticing are passages that are reminiscent of the beginning at bar 13 and at bar 23. 
}
\label{fig:generated}
\end{figure}

Here, we assume that the model is able to learn the temporal dependencies in the musical examples over all timescales and use it to autonomously generate new pieces of music according to those dependencies. For the results presented here, we manually set the first two notes of the melody before generating notes by sampling from the output distributions. This step can be automatized but is needed in order to observe a distribution over possible upcoming notes at the model output. We iteratively add notes to the generated musical sequence until the ``song ending'' output is sampled. 

%REVIEW Why are 2 notes needed?

We observe that the generated melodies (one example is shown in Fig.
\ref{fig:generated}) are different from those found in the training set while
carrying the same features on many time scales. Interestingly, the model is not
explicitly aware of the time signature of each song, but bars can be easily added \emph{a
posteriori} indicating that the long-range rhythmical structure is perfectly
learned and reproduced by our model. The generated melodies are produced according to
production rules extracted from the training set, effectively creating an original composition
that carries the features of the examples in the training
dataset. The model is consequently able to generate new pieces of music
in a completely autonomous manner. 

\section{Discussion}
%%% Short Summary
We fitted a statistical model that uses a recurrent neural network to a
dataset of 2158 Irish melodies.  Due to its recurrent connections and
multiplicative units, the model is able to capture long-range temporal structure. %,as opposed to Markov chain models. 
The model contains no prior knowledge about a musical
style but extracts all relevant features directly from the data. It can
therefore be readily applied to other datasets of different musical styles. For
example, training the model on the Nottingham Music
Database\footnote{http://abc.sourceforge.net/NMD/} yielded similar performance
(data not shown).

%REVIEW: How will your approach generalise with different training sets and or considering different music styles?

%We trained a RNN on a large dataset of Irish folk songs to test the ability of the model to learn both the short-range and the long-range temporal dependencies of music.
We studied the model in two different settings. As a tool for composers, it can
provide song continuations that are coherent with the beginning of the song both
in terms of pitches and in terms of rhythmical patterns. The model also allows
the autonomous composition of new and complete musical sequences. The generated songs
exhibit coherent metrical structure, in some cases temporary modulations to
related keys, and are in general pleasant to hear.

%In future studies, we want to formaly evaluate the generated musical sequences. Indeed, while we used a quantitative measure from statistics (likelihood) in order to train our model and qualitative measures from musical theory, we lack a systematic measure to evaluate the generated melodies from a musical point of view. We want to develop evaluations protocols inspired by the imitation gamed from Turing and use creativness evaluation tools such as the SPECS framework \cite{Jordanous2012}.

%REVIEW:A major issue with the paper is the lack of a formal evaluation of the proposed system. A few generated melodies are included and commented, but this is a qualitative evaluation only. A quantitative assessment ? based for example on the SPECS framework (Jordanous, 2013) ? would certainly improve the quality of the paper.
%%% Relation to other work
% ML approach to algorithmic composition

% Curse of dimensionality in n-grams

% Trainability of neural nets in contrast to stochastic grammars
Using RNNs for algorithmic composition allows to
overcome the limitations of Markov chains in learning the long-range temporal
dependencies of music \cite{Todd1989}.  A different class of models, namely
artificial grammars, are naturally suited to generate these long-range
dependencies due to their hierarchical structure.  However, artificial grammars
have proven to be much harder to learn from data than RNNs
\cite{Gold1967}. Therefore, researchers and composers usually define their own
production rules in order to generate music \cite{Fernandez2013}.  Attempts have
been made to infer context-free grammars
\cite{Nevill-Manning1997} but applications to music are restricted to simple
cases \cite{Kitani2010}. 
%\footnote{The work should be clearly positioned with regards to existing systems exploiting AI approaches for automated melodies generation. At least, other families of approaches should be mentioned and compared.}.}

Evolutionary algorithms constitute another class of approaches that has been very popular in algorithmic composition \cite{Fernandez2013}.
They require the definition of a fitness function, i.e. a measure of the quality of musical compositions (the \textit{individuals}, in the context of evolutionary algorithms). 
Based on the fitness function, the generative process corresponding to the best individuals is favoured and can undergo random mutations.
This type of optimization process is not convenient for generative models for which gradient information is available, like neural networks. However, it can be used in rule-based generative models for which training is hard \cite{Dahlstedt2007}.
A common problem with evolutionary algorithms in music composition is that the definition of the fitness function is arbitrary and requires some kind of evaluation of the musical quality. This resulted in fitness functions often very specific to a certain style.
However, similarly to neural networks, fitness functions can be defined based on the statistical similarity of the individuals to a target dataset of compositions \cite{Alfonseca2005}.
% very often changing from author to author, and therefore not well suited to generalize to different styles. 
%Approaches based on statistical properties of the population of candidate compositions and on their distance from a target set have been proposed \cite{Alfonseca2005, Dahlstedt2007}.
%An evolutionary algorithm whose fitness function is based on statistical distance between the population of composition and whose genomes are the parameters of a generative model would be quite similar to our approach. It would indeed provide a different optimization procedure for the same problem.
%These approaches are more similar to our model, effectively providing a different optimization  evolutionary algorithms usually do not result in a generative model of music but only in a set of composition.
%In contrast, with our approach and with RNNs in general, after the generative model is learned we can use it to generate as many compositions as we want with very low computational cost.

Because of the ease of fitting these models to data as well as their expressiveness,
learning algorithmic composition with recurrent neural networks seems promising.
However, further quantitative evaluations are desirable. We see two quite different
approaches for further evaluation. First, following standard practice in machine
learning, our model could be compared to other approaches in terms of
generalization, which is measured as the conditional probability of held-out
test data (songs) of similar style. Second, the generated songs could be
evaluated by humans in a Turing test setting or with a questionnaire following
the SPECS methodology \cite{Jordanous2012}. Another direction for interesting
future work would be to fit a model to a corpus of polyphonic music and examining the influence of different representations.

\subsubsection*{Acknowledgments} 

Research was supported by the Swiss National Science Foundation (200020\_147200) and the European Research Council grant no. 268689 (MultiRules).

%References: usually there are 10~15 references are ideal in case of conference paper. Depending on requirement the number may change.
\bibliographystyle{unsrt}

\end{document}